\documentclass[runningheads]{llncs}
\usepackage{graphicx}
\usepackage{multirow}
\begin{document}
\title{Deep Learning for Automatic Tracking of Tongue Surface in Real-time Ultrasound Videos, Landmarks instead of Contours}
\titlerunning{TongueNet for Ultrasound Tongue Landmark Tracking}
\author{M. Hamed Mozaffari, Won-Sook Lee}
\institute{School of Electrical Engineering and Computer Science, University of Ottawa, Ottawa, Canada, {mmoza102, wslee}@uottawa.ca}
\maketitle              
\begin{abstract}
One usage of medical ultrasound imaging is to visualize and characterize human tongue shape and motion during a real-time speech to study healthy or impaired speech production. Due to the low-contrast characteristic and noisy nature of ultrasound images, it might require expertise for non-expert users to recognize tongue gestures in applications such as visual training of a second language. Moreover, quantitative analysis of tongue motion needs the tongue dorsum contour to be extracted, tracked, and visualized. Manual tongue contour extraction is a cumbersome, subjective, and error-prone task. Furthermore, it is not a feasible solution for real-time applications.
\\
The growth of deep learning has been vigorously exploited in various computer vision tasks, including ultrasound tongue contour tracking. In the current methods, the process of tongue contour extraction comprises two steps of image segmentation and post-processing. This paper presents a new novel approach of automatic and real-time tongue contour tracking using deep neural networks. In the proposed method, instead of the two-step procedure, landmarks of the tongue surface are tracked. This novel idea enables researchers in this filed to benefits from available previously annotated databases to achieve high accuracy results. Our experiment disclosed the outstanding performances of the proposed technique in terms of generalization, performance, and accuracy.

\end{abstract}
\section{Introduction}
Ultrasound technology is a widespread technology in speech research for studying tongue movement and speech articulation \cite{fasel2010deep} due to its attractive characteristics, such as imaging at a reasonably rapid frame rate, which empowers researchers to envision subtle and swift gestures of the tongue in real-time. Besides, ultrasound technology is portable, relatively affordable, and clinically safe and non-invasive \cite{stone2005guide}. The mid-sagittal view is regularly adapted in ultrasound data as it displays relative backness, height, and the slope of various areas of the tongue. Quantitative analysis of tongue motion needs the tongue contour to be extracted, tracked, and visualized. 
\\
Manual frame-by-frame tongue contour extraction is a cumbersome, subjective, and error-prone task. Moreover, it is not a feasible solution for real-time applications. In conventional techniques, for extracting ultrasound tongue contours, a discrete set of vertices are first annotated near the lower part of the tongue dorsum defining initial deformable tongue contour \cite{xu2016comparative}. Then, through an iterative minimization process using features of the image, the annotated contour is regulated toward the tongue dorsum region. For instance, in active contour models technique (e.g., EdgeTrak software) \cite{li2005automatic}, \cite{xu2016robust}, two internal and external energy functions are minimized over the image gradient. The requirement of feature extraction for each image and accurate initialization are two main drawbacks for those classical techniques. 
Another alternative scenario is to use semi-supervised machine learning models for automatic segmentation of tongue contour regions. Then, tongue contours are extracted automatically using post-processing stages. 
\\
Semi-supervised machine learning-based methods \cite{laporte2018multi} are first utilized for ultrasound tongue contour segmentation in an study by \cite{berry2011dynamics}, \cite{tang2012tongue} while deep learning models emerge in this field through studies by \cite{fasel2010deep}, \cite{jaumard2016tongue}. They fine-tuned one pre-trained decoder part of a Deep Belief Network (DBN) model to infer tongue contours from new instances. End-to-end fashion supervised deep learning techniques, outperformed previous techniques in recent years. For example, U-net \cite{ronneberger2015u} has been used for automatic ultrasound tongue extraction \cite{zhu2019cnn}, \cite{mozaffari2018guided}. After successful results of deep learning methods, the focus of advanced techniques for tongue contour extraction is more on generalization and real-time performance \cite{hamed2019domain}, \cite{mozaffari2019ultrasound}, \cite{mozaffari2019irisnet}. 
\\
Although deep learning methods have been utilized successfully in many studies, manual annotation of ultrasound tongue databases is still cumbersome, and the performance of supervised methods mostly depends on the accuracy of the annotated database as well as the number of available samples. Available databases in this field are annotated by linguistics experts for many years employing landmark points on the tongue contours. In this work, we proposed a new direction for the problem of ultrasound tongue contour extraction using a deep learning technique where instead of tracking the tongue surface, landmarks on the tongue are tracked. In this way, researchers can use previously available linguistics ultrasound tongue databases. Moreover, the whole process of tongue contour extraction is performed in one step, where it increases the performance speed without comprising accuracy or generalization ability of the previous techniques.

\section{Methodology}

\begin{figure}[!thpb]
\centering
\includegraphics[width=\textwidth]{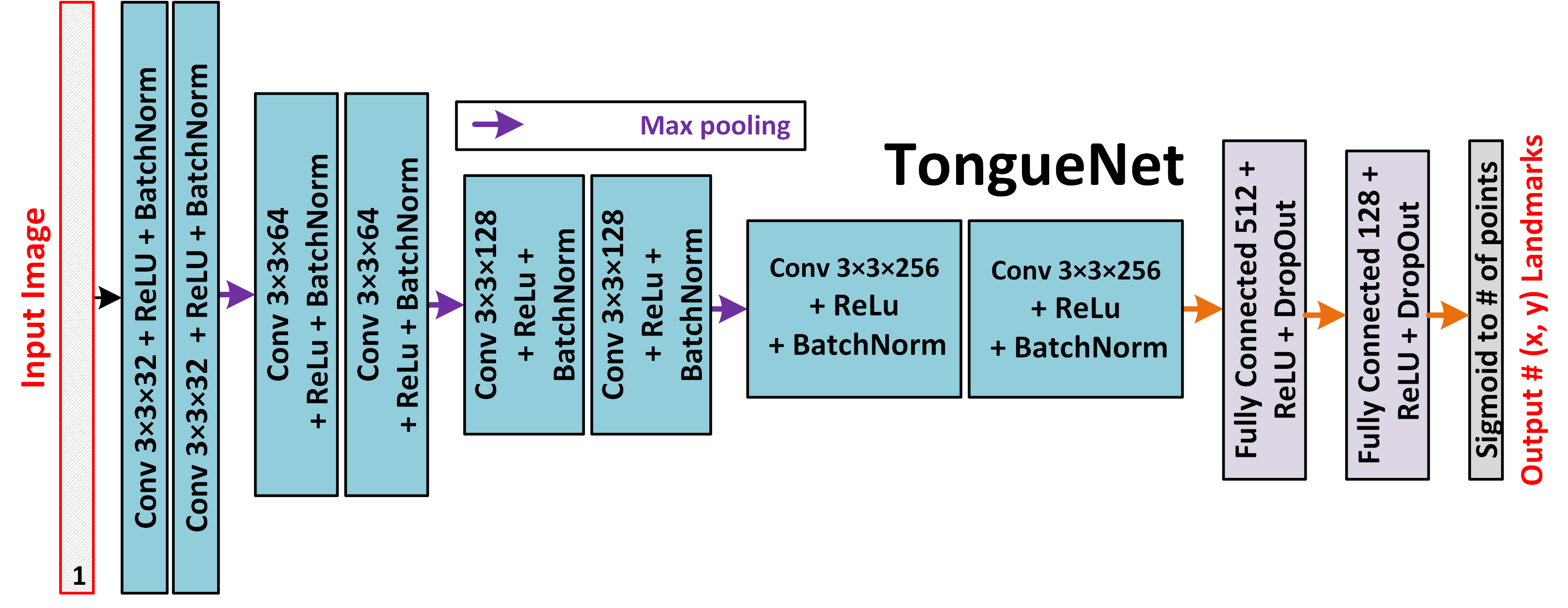}
\caption{An overview of network architecture. Output of the network is a vector comprises of spacial location of individual points on the tongue surface, where $\#$ indicates the number of points in the output }
\label{fig1}
\end{figure}

\begin{figure}[!thpb]
\centering
\includegraphics[width=0.80\textwidth]{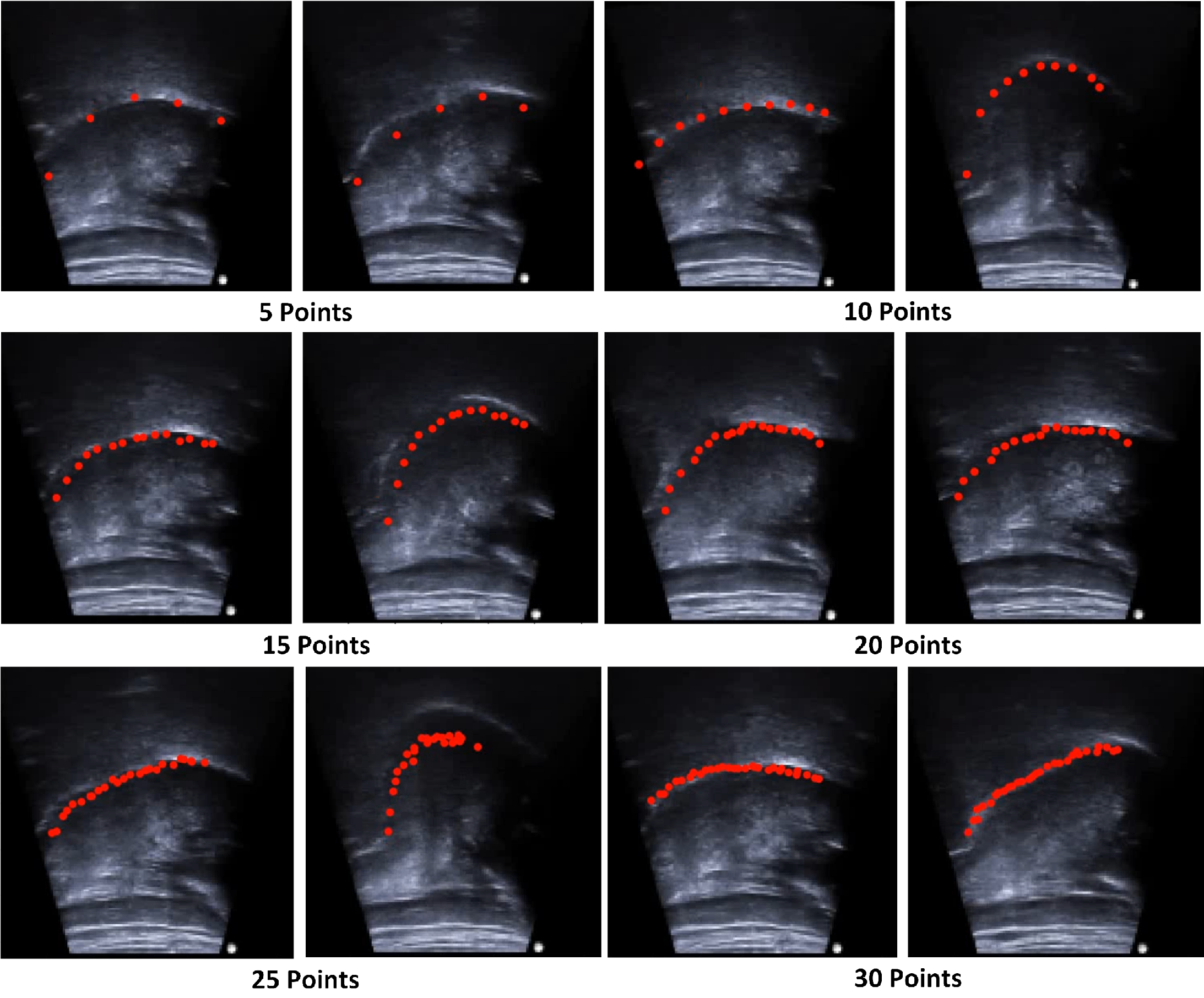}
\caption{Sample frames from the experiment of testing different number of points in the output of the TongueNet.}
\label{fig2}
\end{figure}

Similar to facial landmark detection methods \cite{zhang2014facial}, we considered the problem of tongue contour extraction as a simple landmark detection and tracking. For this reason, we first developed a customized annotator software that can extract equally separated and randomized markers from segmented tongue contours in different databases. Meanwhile, the same software could fit B-spline curves on the extracted markers to revert the process for evaluation purposes. 
\\
To track landmarks on the tongue surface, we designed a light-version deep convolutional neural network named TongueNet. Figure \ref{fig1} shows TongueNet architecture. In each layer, convolutional operations followed by ReLU activations as a non-linearity function as well as Batch normalization layers to improve the regularization, convergence, and accuracy. For the sake of better generalization ability of the model, in the last two layers, fully connected layers are equipped with Drop-out layers of $50\%$. To find the optimum number of required points in the output layer, we used the number of points ($\#$ in Figure \ref{fig1}) from 5 to 100 (see Figure \ref{fig2} for samples of this experiment with 5, 10, 15, 20, 25, and 30 points as the output). 

\section{Experimental Results and Discussion}

\begin{figure}[!thpb]
\centering
\includegraphics[width=\textwidth]{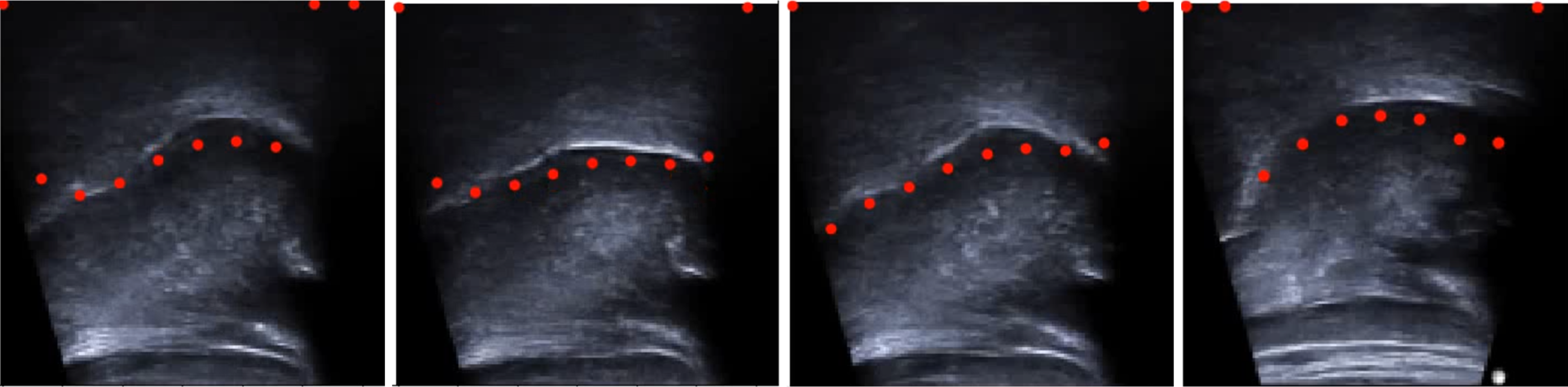}
\caption{Sample results from testing images that were annotated by points equally spaced on the tongue surface through the width of the image. In this image we separated the space by 10 equally disperse vertical lines.}
\label{fig3}
\end{figure}

\begin{figure}[!thpb]
\centering
\includegraphics[width=\textwidth]{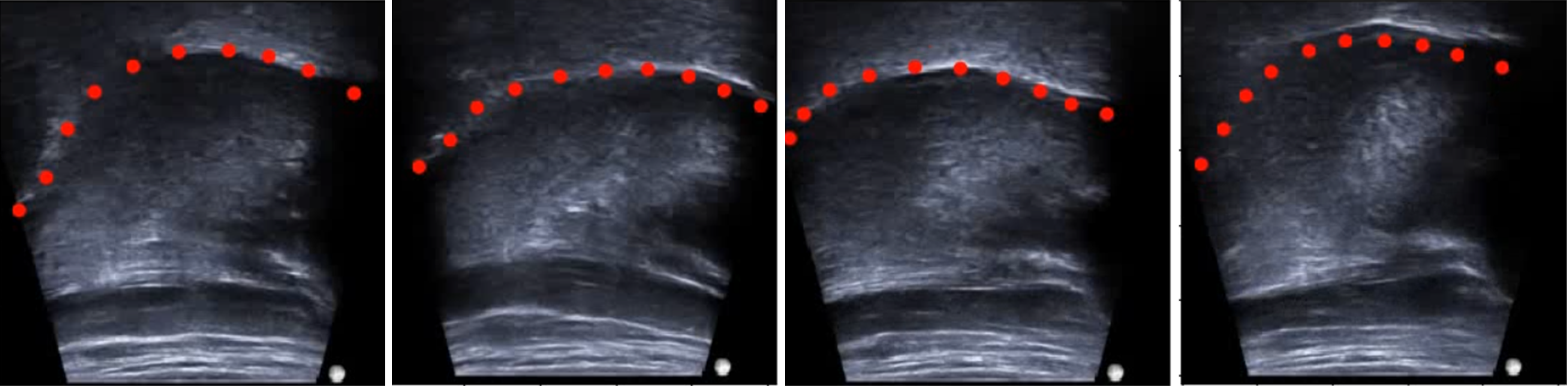}
\caption{Instances generated by TongueNet using randomly selected landmarks on tongue surface, automatically.}
\label{fig4}
\end{figure}

There is usually a trade-off between the number of training samples and the number of trainable parameters in a deep network model \cite{badrinarayanan2017segnet}. In general, more data we have, better result are generated by supervised deep learning methods. Data augmentation helps to increase the number of training data, but a bigger dataset needs a better and most likely bigger network architecture in terms of generalization. Otherwise, the model might over-fitted or under-fitted on training data. Using our annotation software, we automatically extracted landmarks of 2000 images from the UOttawa database \cite{mozaffari2019irisnet} had been annotated for image segmentation tasks. The database was randomly divided into three sets: 90$\%$ training, 5$\%$ validation, and 5$\%$ testing datasets. For testing, we also applied the TongueNet on the UBC database \cite{mozaffari2019irisnet} without any training to see the generalization ability of the model. During the training process of TongueNet, we employed an online data augmentation, including rotation (-25 to 25 degrees), translation (-30 to 30 points in two directions), scaling (from 0.5 to 2 times), horizontal flipping, and combination of these transformations, and annotation point locations are also was transformed, correspondingly. From our extensive random search hyper-parameter tuning, learning rate, the number of iterations, mini-batch sizes, the number of epochs was determined as 0.0005, 1000, 30, 10, respectively. We deployed our experiments using Keras with TensorFlow as the backend on a Windows PC with Core i7, 4.2 GHz speed using one NVIDIA 1080 GPU unit, and 32 GB of RAM. Adam optimization with fixed momentum values of 0.9, was utilized for training. 
\\
We trained and tested TongueNet with a different number of points as the output size. We first evaluated the scenario of equally spaced landmarks on the tongue surface. In this method, we captured all the points in equal distances respect to their neighbors. Our results from selecting five points to the number of pixels in the horizontal axis of the image (image width) revealed that the results of equally spaced selected points are not significant. Figure \ref{fig3} shows randomly selected frames from real-time tracking of TongueNet using ultrasound tongue landmarks. As can be seen from the figure, automatically selection of points as annotation, which is used in many previous studies (see \cite{csapo2015error} for an example), can not provide accurate results. In a separate study, we extract annotation points from the same database using a randomly spaced selection method on tongue contours. We add restrictions for points that they should be at a minimum distance from each other as well as omitting outliers from the database. We saw that the optimum number of points was ten points for our experiments. Figure \ref{fig4} shows some randomly selected results from training TongueNet on a randomly selected point database. From the figure, better accuracy of the TongueNet can be seen qualitatively. Note that we didn't apply any image enhancement or cropping for databases. 

To show the performance ability of TongueNet quantitatively, we first fitted B-spline curves using the OpenCV library on the instances of TongueNet. Then, we compared the value of the mean sum of distance (MSD) \cite{jaumard2016tongue} for TongueNet, sUNET \cite{mozaffari2018guided}, UNET \cite{zhu2018automatic}, BowNet \cite{mozaffari2019ultrasound}, and IrisNet \cite{mozaffari2019irisnet} deep learning models. From table \ref{table1}, it can be seen that TongueNet could reach to results similar to the state of the art deep learning models in this field. Note that there are some approximation errors for the curve-fitting procedure of the TongueNet and skeletonization process for extracting tongue contours from segmentation results of other deep learning models. 
\\
We even tested TongueNet on a new database from UBC (the same database used in \cite{mozaffari2019irisnet}) to evaluate the generalization ability of the landmark tracking technique. Although  TongueNet has not trained on that database, it could predict favorable instances for video frames with different data distributions. This shows the capability of TongueNet for managing of the over-fitting. From Table. \ref{table1}, the difference of MSD values is not significant between models, while IrisNet could find better MSD value. However, in terms of speed, TongueNet outperforms other deep learning models while post-processing time is not considered for them. 

\begin{figure}[!thpb]
\centering
\includegraphics[width=\textwidth]{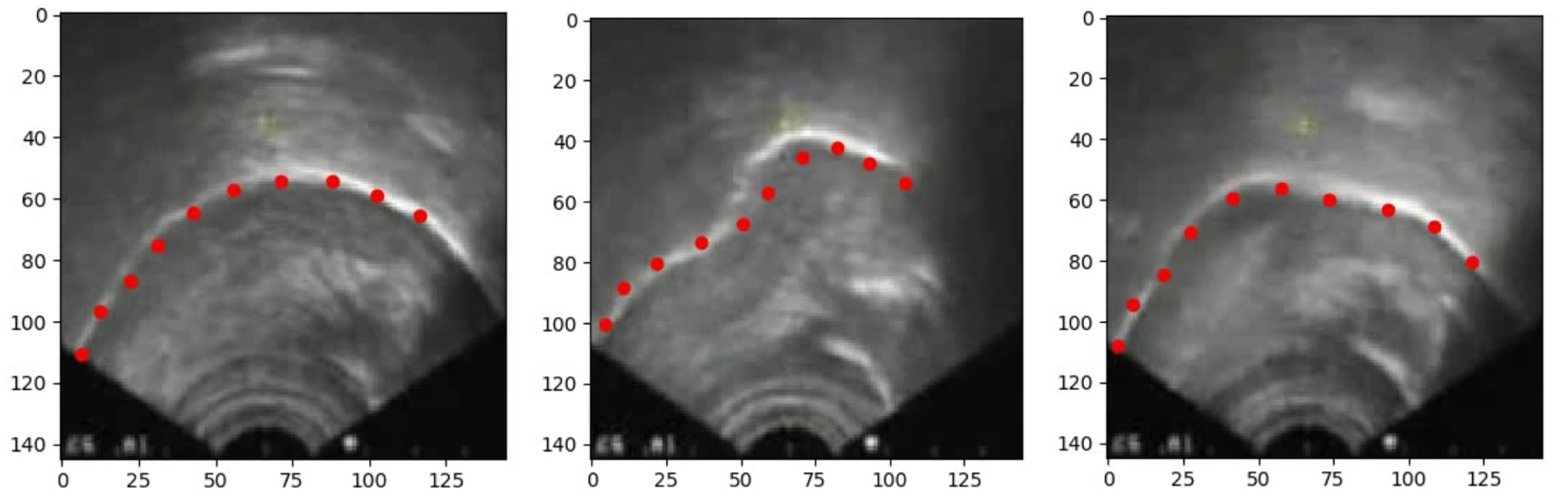}
\caption{Randomly selected frames of applying TongueNet on a new database without training on that.}
\label{fig5}
\end{figure}

We likewise tested TongueNet on a new database from the University of British Columbia (the same database used in \cite{mozaffari2019irisnet}) to evaluate the generalization ability of the landmark tracking technique. Although TongueNet has not trained on that database, it could predict favorable instances for that novel video frames with different data distributions. This shows the capability of TongueNet for managing of the over-fitting. From Table. \ref{table1}, although the difference of MSD values is not significant between models, IrisNet could find better MSD value. However, in terms of speed performance, TongueNet outperforms other deep learning models while post-processing time was not included for calculation of frame rates.

\begin{table}[!thpb]
\centering
\caption{Results of comparison study for several deep learning models on the same database. Except for TongueNet, in order to calculate MSD values, tongue contours were extracted from segmented instances using post-processing method.}
\begin{tabular}{|l|l|l|l|l|l|}
\hline
             & TongueNet & sUNET & UNET & BowNet & IrisNet \\ \hline
MSD (pixels) & 4.87      & 6.81  & 5.32 & 6.12   & 4.21    \\ \hline
FrameRate (f/s)  & 67        & 54    & 46   & 38     & 45      \\ \hline
\end{tabular}
\label{table1}
\end{table}

\section{Conclusion and Discussion}

In this paper, we presented TongueNet, a simple deep learning architecture, with a novel training approach for the problem of Tongue contour extraction and tracking. Unlike similar studies, we used several points to track on the tongue surface instead of the whole tongue contour region. In recent tongue contour tracking, for quantitative studies, a two-phase procedure is performed, including image segmentation and post-processing for extraction of tongue contour that this increase computational costs. Furthermore, available previously annotated tongue databases in articulation studies could not be utilized for deep learning techniques until now. Using TongueNet, we provided a new tool for this literature, while old databases now can be used for developing a better tongue contour tracking technique. 
\\ 
All landmark locations in this study are annotated automatically with two different approaches, and we experienced that data augmentation has a significant rule in the accuracy of TongueNet. From our experimental results, we can anticipate that if an extensive manually annotated database, might be a combination of several databases, is employed for training of a deep learning model such as TongueNet, the accuracy of the model would be boosted considerably. The materials of this study will help researchers in different fields such as linguistics to study tongue gestures in real-time easier, accessible, and with higher accuracy than previous methods. The current infant TongueNet technique needs to be developed, trained, and extended as a fast, accurate, real-time, automatic method applicable for available ultrasound tongue databases. 

\bibliographystyle{splncs04}

\begin{thebibliography}{10}
\providecommand{\url}[1]{\texttt{#1}}
\providecommand{\urlprefix}{URL }
\providecommand{\doi}[1]{https://doi.org/#1}

\bibitem{badrinarayanan2017segnet}
Badrinarayanan, V., Kendall, A., Cipolla, R.: Segnet: A deep convolutional
  encoder-decoder architecture for image segmentation. IEEE transactions on
  pattern analysis and machine intelligence  \textbf{39}(12),  2481--2495
  (2017)

\bibitem{berry2011dynamics}
Berry, J., Fasel, I.: Dynamics of tongue gestures extracted automatically from
  ultrasound. In: 2011 IEEE ICASSP. pp. 557--560. IEEE (2011)

\bibitem{csapo2015error}
Csap{\'o}, T.G., Lulich, S.M.: Error analysis of extracted tongue contours from
  2d ultrasound images. In: Sixteenth Annual Conference of the International
  Speech Communication Association (2015)

\bibitem{fasel2010deep}
Fasel, I., Berry, J.: Deep belief networks for real-time extraction of tongue
  contours from ultrasound during speech. In: 2010 20th ICPR. pp. 1493--1496.
  IEEE (2010)

\bibitem{hamed2019domain}
Hamed~Mozaffari, M., Lee, W.S.: Domain adaptation for ultrasound tongue contour
  extraction using transfer learning: A deep learning approach. The Journal of
  the Acoustical Society of America  \textbf{146}(5),  EL431--EL437 (2019)

\bibitem{jaumard2016tongue}
Jaumard-Hakoun, A., Xu, K., others.: Tongue contour extraction from ultrasound
  images based on deep neural network. In: The International Congress of
  Phonetic Sciences (2015)

\bibitem{laporte2018multi}
Laporte, C., M{\'e}nard, L.: Multi-hypothesis tracking of the tongue surface in
  ultrasound video recordings of normal and impaired speech. Medical image
  analysis  \textbf{44},  98--114 (2018)

\bibitem{li2005automatic}
Li, M., Kambhamettu, C., Stone, M.: Automatic contour tracking in ultrasound
  images. Clinical linguistics \& phonetics  \textbf{19}(6-7),  545--554 (2005)

\bibitem{mozaffari2018guided}
Mozaffari, M.H., Guan, S., Wen, S., Wang, N., Lee, W.S.: Guided learning of
  pronunciation by visualizing tongue articulation in ultrasound image
  sequences. In: 2018 IEEE (CIVEMSA). pp.~1--5. IEEE (2018)

\bibitem{mozaffari2019ultrasound}
Mozaffari, M.H., Kim, C., Lee, W.S.: Ultrasound tongue contour extraction using
  dilated convolutional neural network. In: 2019 IEEE International Conference
  on Bioinformatics and Biomedicine (BIBM). pp. 707--710. IEEE (2019)

\bibitem{mozaffari2019irisnet}
Mozaffari, M.H., Ratul, M., Rab, A., Lee, W.S.: Irisnet: Deep learning for
  automatic and real-time tongue contour tracking in ultrasound video data
  using peripheral vision. arXiv preprint arXiv:1911.03972  (2019)

\bibitem{ronneberger2015u}
Ronneberger, O., Fischer, P., Brox, T.: U-net: Convolutional networks for
  biomedical image segmentation. In: International Conference on MICCAI. pp.
  234--241. Springer (2015)

\bibitem{stone2005guide}
Stone, M.: A guide to analysing tongue motion from ultrasound images. Clinical
  linguistics \& phonetics  \textbf{19}(6-7),  455--501 (2005)

\bibitem{tang2012tongue}
Tang, L., Bressmann, T., Hamarneh, G.: Tongue contour tracking in dynamic
  ultrasound via higher-order mrfs and efficient fusion moves. Medical image
  analysis  \textbf{16}(8),  1503--1520 (2012)

\bibitem{xu2016comparative}
Xu, K., G{\'a}bor~Csap{\'o}, T., Roussel, P., Denby, B.: A comparative study on
  the contour tracking algorithms in ultrasound tongue images with automatic
  re-initialization. The Journal of the Acoustical Society of America
  \textbf{139}(5),  EL154--EL160 (2016)

\bibitem{xu2016robust}
Xu, K., Yang, Y., others.: Robust contour tracking in ultrasound tongue image
  sequences. Clinical linguistics \& phonetics  \textbf{30}(3-5),  313--327
  (2016)

\bibitem{zhang2014facial}
Zhang, Z., Luo, P., Loy, C.C., Tang, X.: Facial landmark detection by deep
  multi-task learning. In: European conference on computer vision. pp. 94--108.
  Springer (2014)

\bibitem{zhu2019cnn}
Zhu, J., Styler, W., Calloway, I.: A cnn-based tool for automatic tongue
  contour tracking in ultrasound images. arXiv preprint arXiv:1907.10210
  (2019)

\bibitem{zhu2018automatic}
Zhu, J., Styler, W., Calloway, I.C.: Automatic tongue contour extraction in
  ultrasound images with convolutional neural networks. The Journal of the
  Acoustical Society of America (Poster)  \textbf{143}(3),  1966--1966 (2018)

\end{thebibliography}

\end{document}